# An Improved BAT Algorithm for Solving Job Scheduling Problems in Hotels and Restaurants

**Tarik A. Rashid[1*], Chra I. Shekho Toghramchi[1], Heja Sindi[2], Abeer Alsadoon[3], Nebojsa Bacanin[4], Shahla U. Umar[5,6], A.S. Shamsaldin[1], Mokhtar Mohammadi[7]**

[1]Department of Computer Science and Engineering, University of Kurdistan of Hewler, Erbil, KRG, Iraq
[2]Business and Management, School of Management and Economics, University of Kurdistan Hewlêr, Erbil, KRG, Iraq.
[3]Charles Sturt University, Sydney, Australia
[4]Singidunum University, Belgrade, Serbia
[5]Technical College of Informatics, Sulaimani Polytechnic University, Sulaimani, KRG, Iraq
[6]Network Department, College of Computer Science and Information Technology, Kirkuk University, Kirkuk, Iraq.
[7]Department of Information Technology, Lebanese French University, Erbil, 44001, KRG, Iraq;

*Corresponding author's email: Tarik.ahmed@ukh.edu.krd

**Abstract:** One popular example of metaheuristic algorithms from the swarm intelligence family is the Bat algorithm (BA). The algorithm was first presented in 2010 by Yang and quickly demonstrated its efficiency in comparison with other common algorithms. The BA is based on echolocation in bats. The BA uses automatic zooming to strike a balance between exploration and exploitation by imitating the deviations of the bat's pulse emission rate and loudness as it searches for prey. The BA maintains solution diversity using the frequency-tuning technique. In this way, the BA can quickly and efficiently switch from exploration to exploitation. Therefore, it becomes an efficient optimizer for any application when a quick solution is needed. In this paper, an improvement on the original BA has been made to speed up convergence and make the method more practical for large applications. To conduct a comprehensive comparative analysis between the original BA, the modified BA proposed in this paper, and other state-of-the-art bio-inspired metaheuristics, the performance of both approaches is evaluated on a standard set of 23 (unimodal, multimodal, and fixed-dimension multimodal) benchmark functions. Afterward, the modified BA was applied to solve a real-world job scheduling problem in hotels and restaurants. Based on the achieved performance metrics, the proposed MBA establishes better global search ability and convergence than the original BA and other approaches.

## 1. Introduction

Swarm intelligence (SI) is described as the collective behaviour of self-organized and distributed systems that might be artificial or natural. It was introduced in 1989, by Beny [1]. Since then, a great number of algorithms have been proposed. Swarm intelligence algorithms substantiated their ability to tackle numerous optimization problems in several areas. Self-organization and the separation of work are the essential characteristics of swarm intelligence. Self-organization means the ability to organize any system into a preferred pattern or model without any external support. Self-organization also depends on several characteristics, i.e., multiple interactions, negative feedback, fluctuations, and positive feedback. To achieve amplification and stabilization, positive and negative feedback are implemented, while fluctuations are useful for randomness. Meanwhile, different interactions take place when swarm individuals share information among themselves in the search space [2].

Advanced optimization algorithms based on swarm intelligence are inspired by nature. The main sources are biological systems, such as Ant Colony Optimization(ACO), the Genetic Algorithm (GA), the Cuckoo Search Algorithm (CS), the Bat Algorithm, and many others [2]. Nowadays, nature-inspired meta-heuristic algorithms have become a useful tool for finding the answers to optimization problems that companies encounter. In meta-heuristic algorithms, there are two types of solutions. One is single-solution-based and the other is population-based. In single solution-based algorithms, the search process begins with only a single candidate solution, which is then improved over the iterations of the algorithm. In contrast, agents in the population-based one initialize random population in the first iteration and have better communication. Possible solutions exchange information about the search space, which leads to the search process moving to more feasible areas and avoiding being trapped in local optima, Furthermore, it has better exploration ability [3], [4].

The Bat Algorithm is characterized by its efficiency in the exploitation phase, while it suffers from a weak exploration capacity. Sometimes bats get outside the solution search space. Also, acceptance of the scheme is one of the limitations of the BA's need to reduce the probability of accepting a poor solution. In practice, it shows that the BA converges very quickly in the early stages, while it later slows down. However, the accuracy is sufficient when the number of functions is high. The BA does not save the best loudness in every single iteration, even though a global solution has very little impact on a new solution. The main aim of this work is to present a modified version of the BA, then test it against 23 benchmark functions. The results of the BA are then compared to those of the MBA. Finally, a real-world job scheduling problem is solved using the MBA. The main contribution of the MBA is to increase the convergence level and improve the accuracy of the standard BA.

The rest of the paper is presented as follows: in section two, a literature review is presented. In section three, details of the Bat algorithm are explained. Section four introduces the MBA. Section five presents the implementation of both the BA and the MBA. Finally, in section 6, the results are provided along with an analysis.

## 2. Literature Review

In 2013, Ramesh, Mohan, and Reddy presented an application that aimed to eliminate the environmental effects caused by fossil-fuel power plant emissions. The application used a Bat algorithm to deal with a multi-objective optimization problem in the power system.

[5].Yilmaz, Kucuksille, and Cengiz proposed an improvement to the exploration mechanism of the Bat algorithm in 2014. They modified the loudness and pulse rate emission of the Bat algorithm since these two factors affect all dimensions in the solutions while searching. They checked the performance of the modified Bat algorithm and BA on 15 benchmark test functions. When it comes to solving optimization problems, the MBA outperformed the original BA. [6].

In 2015, Fister, Rauter, Yang, Ljubic, and Fister Jr, proposed an MBA as an intelligent planner for sports training sessions. They modified the Bat algorithm to solve the mentioned problem. They applied the MBA algorithm on ten basic training sessions for a specific athlete. The idea was to plan a cycling training session The results showed a plan whose prediction complied with the high standards required by cycling coaches [7].

In 2016, Kielkowics and Grela proposed two modifications of the bat algorithm. They introduced a different scheme of acceptance for newly found solutions and the velocity equation was modified. Furthermore, they modified the acceptance scheme to reduce the probability of accepting a poor solution, and also changed the velocity update equation by archiving the component and introducing a cognitive coefficient. As the last step, they created additional memory by storing the best solutions that were found during the optimization process. They compared the modified bat algorithm with the existing bat algorithm and tested the modifications in a few simulation experiments [8].

Cai, Gao, and Xue introduced an improved Bat algorithm in 2016. The improvements focus on the exploitation and exploration capabilities of the Bat algorithm, which was limited, and the multi-model numerical problem. A new search mechanism was suggested using an optimal forage approach to improve the bat's movements and directions, as well as to utilize the random disturbance method to enhance the exploration phase. They tested and compared the performance with four other evolutionary algorithms. The simulation results indicated that the improved Bat algorithm was the most effective[9].

In 2019, Osab et al. proposed a discrete and improved Bat algorithm (DaIBA) that concentrated on the problem of real-world drugs spreading with pharmacological waste collection. They modeled the problem as a multi-attribute, or rich vehicle routing problem (RVRP), i.e. as a clustered vehicle routing problem for deliveries and pickups, asymmetric variable costs, and forbidden paths. This is the first time that the problem was addressed by the scientific community. Thereupon, they designed a benchmark consisting of 24 datasets that contained from 60 to 1000 customers. The results concluded that the proposed algorithm was a promising technique for the specified problem [10].

## 3. The Bat Algorithm

There are some rules which are used to develop Bat-inspired algorithms [11];

1. Distance: In general, bats use echolocation to sense distance. They recognize the difference between food/prey and background barriers in a remarkable way.
2. Frequency: They can fly arbitrarily with a velocity of $v_i$ position $x_i$ with a fixed frequency $f_{min}$, varying wavelength $\gamma$, and loudness $A_0$ to search for prey. They can automatically adjust the wavelength (or frequency) of their emitted pulses and adjust their rate of pulse emission r ∈ [0, 1], depending on their proximity to their target.
3. Loudness: Although the loudness can vary in many ways, it can be assumed that loudness could diverge from a significant (positive) $A_0$ to a minimum constant $A_{min}$.

Generally, $(x_i)$ was used to represent the bat's position in the search space, while the $(v_i)$ and $(f_i)$ used to denote the velocities and the frequency of the pulse respectively. Whereas $(\beta)$ refers to a vector of random numbers between 0 and 1. As well as, $(x_*)$ indicates the best solution obtained till now.

$$f_i = f_{min} + (f_{max} - f_{min})\beta \tag{1}$$

$$v_i^t = v_i^{t-1} + (x_i^t - x_*)f_i \tag{2}$$

$$x_i^t = x_i^{t-1} + v_i^t \tag{3}$$

After comparing all the other solutions of *n* bats. Each bat is assigned a random frequency from $[f_{max}, f_{min}]$. Using a random walk for the local search, only one solution is selected from the best ones at this moment and a new solution for each bat is generated.

$$x_{new} = x_{old} + \epsilon A^t \tag{4}$$

$A^t$, which is equal to $<A_i^t>$ and is the average loudness at this time step for all of the bats and $\epsilon \in$ [-1, 1] is a random number.

As the iteration proceeds, the emission must be updated, as must the loudness $A_i$ and the rate $r_i$ of the pulse. When a bat finds its prey, the loudness decreases, and the pulse emission increases

$$A_i^t = \alpha A_i^t, \qquad r_i^{t+1} = r_i^0[1 - \exp(-\gamma t)] \tag{5}$$

$$\text{If } (rand < A^i \, \& \, f(x_i) < f(x_*)) \tag{6}$$

Where $\alpha$ and $\gamma$ are constants, for any $0 < \alpha < 1$ and $\gamma > 0$,

$$A_i^t \to 0, \, r_i^t \to r_i^0, \, as \, t \to \infty \tag{7}$$

For the sake of simplicity, $\alpha = \gamma$ is used. Through the randomization process, different values for the loudness and pulse emission rate of each bat [11] are computed. Figure 1, shows the pseudo-code of the Bat Algorithm.

**Pseudo Code of Bat Algorithm**
---

Objective Function $f(x)$, $x = (x_1, \ldots, x_d)^T$
Initialize the bat population $x_i$ $(i = 1,2,\ldots,n)$ and $v_i$
Define pulse frequency $f_i$ at $x_i$
Initialize pulse rates $r_i$ and the loudness $A_i$
**while** (t < max number of iterations)
    Generate new solution by adjusting, updating velocities and locations use equation (1), (2), and (3)
    **if** (rand > $r_i$)
        chose a solution among the best solutions
        $x_{new} = x_{old} + \epsilon A^t$
    **end if**
    Generate a new solution by flying randomly
    **If** (rand < $A^t$ & $f(x_i) < f(x_*)$)
        Accept the new solutions
        Increase $r_i$ and decrease $A_i$
    **end if**
    Rank the bats and find the current best $x_*$
**end while**

**Figure 1: shows the pseudo-code of the Bat Algorithm.**

## 4. The Modified Bat Algorithm

In a meta-heuristic-based population, one of the essential aspects is the balance between the exploration and exploitation phases of the search space. Exploration is sometimes called diversification and is responsible for the global search ability of the algorithm. Exploitation is sometimes called intensification and is responsible for the local search ability of the algorithm. The bat algorithm is more effective at exploitation in a local search; although it may become trapped in local optima which leads to an inability to perform a global search [8]. Generally, the BA depends entirely on random walks, therefore swift convergence cannot be guaranteed [11]. The bat algorithm is modified to mitigate the aforementioned limitation. The main improvement is achieved by replacing the old position as opposed to adding a global position and global loudness for all bats. This is done by modifying equation (4) to match equation (8).

$$x_{new} = x^* + \epsilon A^t + A^* \qquad (8)$$

The improvement aims to speed up convergence and make the method more suitable for a wider range of applications. If a new solution improves the result, then the loudness and pulse rate will be updated. After generating the new solution, the loudness of the best solution until that point is saved as the best loudness. Figure 2 shows the pseudo-code of the modified Bat Algorithm.

**Pseudo Code of Modified Bat Algorithm**

```
Objective Function  f(x), x = (x_1, ....., x_d)^T
Initialize the bat population  x_i (i = 1,2, ..., n ) and v_i
Define pulse frequency  f_i at x_i
Initialize pulse rates  r_i and the loudness  A_i
while (t < max number of iterations)
    Generate new solution by adjusting, updating velocities and locations use equation (1), (2) and (3)
    if (rand > r_i )
        chose a solution among the best solutions
        x_new = x* + εA^t+A*
    end if
    Generate a new solution by flying randomly
    If (rand < A^t & f(x_i) < f(x*))
        Accept the new solutions
        Store the loudness of the best solution
        Increase r_i and decrease A_i
    end if
    Rank the bats and find the current best x*
end while
```

**Figure 2 shows the pseudo-code of the modified Bat Algorithm.**

**5. Implementation**

The BA and MBA are tested on a set of standard well-known benchmark test functions. A simulation experiment was performed on a MacBook Pro computer, running macOS Sierra version 10.12.6. on Intel Core i7 2.2 GHz with 16 GB of memory. The algorithms are implemented in C++ and compiled with MATLAB R2018a. This paper is divided into several parts, as shown below:

**5.1 Benchmark Test Functions**

To conduct a comprehensive evaluation of the bat algorithm and the modified bat algorithm, and the effects they have on problems, their performance was tested using a large set of standard benchmark functions. These benchmark functions are listed in Tables 1, 2, and 3. In total, there are 23-benchmark functions, which are classified into three categories according to their characteristics, as follows:

**5.1.1 Unimodal Benchmark Functions**

This group consists of functions from F1 – F7 and they are high dimensional. They are relatively easy to solve in comparison to the other groups. Table (1) shows the unimodal benchmark function's name, equation, and the lower and upper bounds used for the function.

| Function Name | Function | Range |
|---|---|---|
| Sphere function | $f_1(x) = \sum_{i=1}^{n} x_i^2$ | [-100, 100] |
| Schwefel's problem 2.22 | $f_2(x) = \sum_{i=1}^{n} |x_i| + \prod_{i=1}^{n} |x_i|$ | [-10, 10] |
| Schwefel problem 1.2 | $f_3(x) = \sum_{i=1}^{n} \left(\sum_{j-1}^{i} x_j\right)^2$ | [-100, 100] |
| Schwefel's problem 2.21 | $f_4(x) = max_i\{|x_i|, \quad 1 \leq i \leq n\}$ | [-100, 100] |

| | | |
|---|---|---|
| Generalized Rosenbrock's function | $f_5(x) = \sum_{i=1}^{n-1} [100(x_{i+1} - x_i^2)^2 + (x_i - 1)^2]$ | [-30, 30] |
| Step function | $f_6(x) = \sum_{i=1}^{n} ([x_i + 0.5])^2$ | [-100, 100] |
| Quartic function with noise | $f_7(x) = \sum_{i=1}^{n} ix_i^4 + random[0,1]$ | [-1.28, 1.28] |

**Table 1: Unimodal benchmark test functions.**

### 5.1.2 Multimodal Benchmark Functions

This group is composed of functions F8- F13. They are considered to be the most challenging problems in the set of benchmark functions. Table (2) shows the Multimodal benchmark function's name, equation, and the lower and upper bounds used for the function.

**Table 2: Multimodal benchmark test functions.**

| Function Name | Function | Range |
|---|---|---|
| Generalized Schwefel's problem 2.26 | $f_8(x) = \sum_{i=1}^{n} -x_i \sin\left(\sqrt{|x_i|}\right)$ | [-500, 500] |
| Generalized Rastrigin's function | $f_9(x) = \sum_{i=1}^{n} [x_i^2 - 10 \cos(2\pi x_i) + 10]$ | [-5.12, 5.12] |
| Ackley's function | $f_{10}(x) = -20 exp\left(-0.2\sqrt{\frac{1}{n}\sum_{i=1}^{n} x_i^2}\right) - exp\left(\frac{1}{n}\sum_{i=1}^{n} \cos(2\pi x_i)\right) + 20 + e$ | [-32, 32] |
| Generalized Griewank function | $f_{11}(x) = \frac{1}{4000} \sum_{i=1}^{n} x_i^2 - \prod_{i=1}^{n} \cos\left(\frac{x_i}{\sqrt{i}}\right) + 1$ | [-600, 600] |
| Generalized penalized function | $f_{12}(x) = \frac{\pi}{n}\{10 \sin(\pi y_1) + \sum_{i=1}^{n-1}(y_i - 1)^2 [1 + 10\sin^2(\pi y_{i+1})] + (y_n - 1)^2\} + \sum_{i=1}^{n} u(x_i, 10, 100, 4)$ $y_i = 1 + \frac{x_i + 1}{4}$ $u(x_i, a, k, m) = \begin{cases} k(x_i - a)^m & x_i > a \\ 0 & -a < x_i < a \\ k(-x_i - a)^m & x_i < -a \end{cases}$ | [-50, 50] |
| Generalized penalized function | $f_{13}(x) = 0.1 \{\sin^2(3\pi x_1) + \sum_{i=1}^{n}(x_i - 1)^2 [1 + \sin^2(3\pi x_i + 1)] + (x_n - 1)^2 [1 + \sin^2(2\pi x_n)]\} + \sum_{i=1}^{n} u(x_i, 10, 100, 4)$ | [-50, 50] |

### 5.1.3 Fixed-Dimension Multimodal Benchmark Functions

This group includes functions F14-F23 and they have smaller dimensions. Table (3) shows the Fixed-Dimension Multimodal benchmark function's name, equation, and the lower and upper bounds used for the function.

**Table 3: Fixed-dimension multimodal benchmark test functions.**

| Function | Function | Range |
|---|---|---|
| Shekel's Foxholes function | $f_{14}(x) = \left(\frac{1}{500} + \sum_{j=1}^{25} \frac{1}{j + \sum_{i=1}^{2}(x_i - a_{ij})^6}\right)^{-1}$ | [-65.536, 65.536] |
| Kowalik's function | $f_{15}(x) = \sum_{i=1}^{11} [a_i - \frac{x_1(b_i^2 + b_i x_2)}{b_i^2 + b_i x_3 + x_4}]^2$ | [-5, 5] |
| Six-hump camel back function | $f_{16}(x) = 4x_1^2 - 2.1x_1^4 + \frac{1}{3}x_1^6 + x_1 x_2 - 4x_2^2 + 4x_2^4$ | [-5, 5] |
| Brain function | $f_{17}(x) = \left(x_2 - \frac{5.1}{4\pi^2}x_1^2 + \frac{5}{\pi}x_1 - 6\right)^2 + 10\left(1 - \frac{1}{8\pi}\right)\cos x_1 + 10$ | [-5,0],[10,15] |
| Goldstein-Price function | $f_{18}(x) = [1 + (x_1 + x_2 + 1)^2(19 - 14x_1 + 3x_1^2 - 14x_2 + 6x_1x_2 + 3x_2^2)]$ $\times [30 + (2x_1 - 3x_2)^2 \times (18 - 32x_1 + 12x_1^2 + 48x_2 - 36x_1x_2 + 27x_2^2)]$ | [-2, 2] |
| Hartman's family | $f_{19}(x) = -\sum_{i=1}^{4} c_i \exp\left(-\sum_{j=1}^{3} a_{ij}(x_j - p_{ij})^2\right)$ | [0, 1] |
| Hartman's family | $f_{20}(x) = -\sum_{i=1}^{4} c_i \exp\left(-\sum_{j=1}^{6} a_{ij}(x_j - p_{ij})^2\right)$ | [0, 1] |
| Shekel's family | $f_{21}(x) = -\sum_{i=1}^{5} [(x - a_i)(x - a_i)^T + c_i]^{-1}$ | [0,10] |
| Shekel's family | $f_{22}(x) = -\sum_{i=1}^{7} [(x - a_i)(x - a_i)^T + c_i]^{-1}$ | [0, 10] |
| Shekel's family | $f_{23}(x) = -\sum_{i=1}^{10} [(x - a_i)(x - a_i)^T + c_i]^{-1}$ | [0,10] |

## 6. Results and Discussion

The bat algorithm and the modified bat algorithm searched for an optimum solution with a population size of 30 and a maximum number of iterations of 500. The goal was to find an optimum solution over 500 iterations after which the average value and standard deviation results of both the BA and the MBA were calculated. Many comparisons were conducted to evaluate the BA and the MBA:

### 6.1 The Average Value and the Standard Deviation

The averages and standard deviation results of the BA and the MBA compared as follows:

1. Unimodal function comparison: In Table (4), for the average value, the MBA had a better result than the BA in all functions, which means that the MBA is a better optimizer. Also, because the unimodal function is an optimum global value, the MBA performed better. At the same time, in Table (5), the standard deviation results showed that the MBA provides better results in 6 benchmark functions (F1, F2, F3, F5, F6, and F7) while the BA is better in only one (F4). This shows the outperforming of the modified bat algorithm against the standard bat.
2. The multimodal function and fixed-dimension multimodal functions: Table (4) illustrates that the MBA was more efficient on average in 13 of the benchmark functions (from F8 until F20), while the BA produced a better optimum value in 3 functions (F21, F22, and F23). These functions have many local optima that depend on the variable design, the number of optima increased. Also for the standard deviation result of the multimodal function and fixed-dimension multimodal functions, Table (5) shows that the MBA produced a better optimum result in 11 of the benchmark functions (F8, F11, F12, F14-F18, and F21-F23), while the BA produced a better optimum value in 5 of the benchmark functions (F9, F10, F13, F19, and F20). The results show that the modified bat algorithm is an improvement on the bat algorithm and that it had better exploration ability.

As the results show, the BA is poor at exploration, which is a weak point for most problems. The MBA helps the BA to be more global and converge on a result more quickly. In the unimodal functions, the MBA obtained a significant result that had global values, this proved that the MBA functions better globally.

**Table 4: Average value comparison.**

| Function | Average Value of BA | Average Value of MBA | Min AVG |
|---|---|---|---|
| F1 | 1.510E+01 | 3.406E+00 | + |
| F2 | 1.898E+01 | 4.096E+00 | + |
| F3 | 9.743E+02 | 7.504E+00 | + |
| F4 | 9.673E-01 | 4.172E-01 | + |

| F5 | 2.638E+03 | 3.038E+02 | + |
| F6 | 2.298E+01 | 3.296E+00 | + |
| F7 | 1.048E+02 | 2.323E+01 | + |
| F8 | -5.498E+01 | -1.164E+02 | + |
| F9 | 2.298E+01 | 1.519E+01 | + |
| F10 | 3.136E+00 | 8.926E-01 | + |
| F11 | 6.362E-01 | 1.307E-01 | + |
| F12 | 2.549E+00 | 3.723E-01 | + |
| F13 | 1.020E+00 | 9.926E-01 | + |
| F14 | 1.268E+01 | 1.267E+01 | + |
| F15 | 1.620E-02 | 1.030E-02 | + |
| F16 | -4.655E-01 | -6.314E-01 | + |
| F17 | 8.158E+00 | 7.072E-01 | + |
| F18 | 2.449E+02 | 4.845E+01 | + |
| F19 | -2.456E+00 | -2.643E+00 | + |
| F20 | -3.643E-01 | -5.514E-01 | + |
| F21 | -2.478E+00 | -2.111E+00 | - |
| F22 | -2.724E+00 | -2.100E+00 | - |
| F23 | -2.756E+00 | -2.494E+00 | - |

**Table 5: Standard deviation comparison.**

| Function | Standard Deviation of BA | Standard Deviation of MBA | Min STD |
|---|---|---|---|
| F1 | 1.139E+01 | 4.258E+00 | + |
| F2 | 8.009E+00 | 5.571E+00 | + |
| F3 | 1.126E+03 | 1.987E+01 | + |
| F4 | 1.555E-01 | 4.340E-01 | - |
| F5 | 7.944E+02 | 4.225E+02 | + |
| F6 | 1.344E+01 | 4.059E+00 | + |
| F7 | 1.227E+02 | 3.008E+01 | + |
| F8 | 1.210E+01 | 2.450E-01 | + |
| F9 | 1.219E+01 | 3.140E+01 | - |
| F10 | 1.179E+00 | 1.488E+00 | - |
| F11 | 2.352E-01 | 2.005E-01 | + |
| F12 | 1.732E+00 | 2.302E-01 | + |
| F13 | 1.537E-01 | 2.264E-01 | - |
| F14 | 4.420E-02 | 3.904E-04 | + |
| F15 | 1.410E-02 | 9.600E-03 | + |
| F16 | 4.749E-01 | 3.491E-01 | + |
| F17 | 2.950E+00 | 3.077E-01 | + |
| F18 | 2.205E+02 | 4.170E+01 | + |
| F19 | 7.736E-01 | 7.862E-01 | - |
| F20 | 4.199E-01 | 5.340E-01 | - |
| F21 | 1.105E+00 | 8.086E-01 | + |
| F22 | 1.068E+00 | 9.243E-01 | + |
| F23 | 1.258E+00 | 9.892E-01 | + |

### 6.2 Statistical Tests

The BA and the MBA results were tested for statistical significance with the Wilcoxon rank-sum test. The t-test is used to determine that the two samples come from the same two populations with the same mean. This t-test was used to calculate the P-Value of the BA and the MBA. The P-Value of the t-test must be lower than 0.05 for results to be statistically significant. As shown in Table (6) 60% percent of the P-Values of the BA and the MBA is below 0.05, which shows that the results of our algorithms are statistically significant.

**Table 6: The Wilcoxon rank-sum test for benchmark functions.**

| Function | BA Vs. MBA (P-Value) |
|---|---|
| F1 | 2.689E-05 |
| F2 | 5.009E-11 |
| F3 | 6.028E-05 |
| F4 | 3.513E-07 |
| F5 | 6.508E-15 |
| F6 | 9.137E-09 |
| F7 | 2.004E-03 |
| F8 | 1.857E-22 |
| F9 | 2.384E-01 |
| F10 | 6.594E-08 |
| F11 | 4.298E-09 |
| F12 | 7.831E-08 |
| F13 | 5.781E-01 |
| F14 | 1.329E-01 |
| F15 | 6.671E-02 |
| F16 | 1.272E-01 |
| F17 | 1.543E-14 |
| F18 | 6.831E-05 |
| F19 | 3.189E-01 |
| F20 | 1.101E-01 |
| F21 | 2.001E-01 |
| F22 | 2.584E-02 |
| F23 | 4.106E-01 |

### 6.3 The MBA vs. the BA

The results shown in this paper are taken from the original paper published by Yang, in which the new Bat algorithm is compared with the Genetic Algorithm (GA) and Particle Swarm Optimization (PSO). This comparison uses the same type of approach (Yang, 2010). In the simulation of the MBA, six functions are being used, i.e. the Generalized Rosenbrock's function, the Spher function, the Ackley's function, the Schwefel problem 1.2, the Generalized Rastrigin's function, and the Generalized Griewank function. The number of populations is set to 40 and the dimension of the algorithm was also changed as follows:

1. The Generalized Rosenbrock's function changes the dimension to (dim=16).
2. De Jong's standard sphere function changes the dimension to (dim=256).
3. The Schwefel problem 1.2 changes the dimension to (dim=128).
4. Ackley's function changes the dimension to (dim=128).

Following the evaluation, which included the MBA, the PSO, and the GA, Table (7) shows a comparison with six functions, which indicates that the MBA's results are better than the PSO and the GA, in all six functions. This proves that the MBA is superior to other algorithms in terms of accuracy and efficiency.

**Table 7: Comparing MBA with PSO and GA**

| Name | Function number | MBA AVG | GA AVG | PSO AVG |
|---|---|---|---|---|
| Generalized Rosenbrock's function | F5 | 2.685E+02 | 5.572E+04 | 3.276E+04 |
| Sphere function | F1 | 2.150E+00 | 2.541E+04 | 1.704E+04 |
| Ackley's function | F10 | 1.176E+00 | 3.272E+04 | 2.341E+04 |
| Schwefel problem 1.2 | F3 | 1.237E+02 | 2.273E+05 | 1.452E+04 |
| Generalized Rastrigin's function | F9 | 2.735E+01 | 1.105E+05 | 7.949E+04 |
| Generalized Griewank function | F11 | 1.077E-01 | 7.093E+04 | 5.597E+04 |

Another comparison involves the MBA and the Genetic Algorithm. The GA data are taken from [12], and used 19 benchmark functions. However, 13 of them are from our list of unimodal and multimodal functions. Table (8), shows the performance of the modified bat algorithm compared with the Genetic Algorithm. The results for the modified bat algorithm were better in 11 functions. The GA was better in only two functions, - one of the unimodal functions and one of the multimodal functions.

Table 8: Comparing MBA and GA

| F | GA AVG | MBA AVG | Min AVG |
|---|---|---|---|
| F1 | 7.486E+02 | 3.406E+00 | + |
| F2 | 5.971E+00 | 4.096E+00 | + |
| F3 | 1.949E+03 | 7.504E+00 | + |
| F4 | 2.116E+01 | 4.172E-01 | + |
| F5 | 1.333E+05 | 3.038E+02 | + |
| F6 | 5.639E+02 | 3.296E+00 | + |
| F7 | 1.669E-01 | 2.323E+01 | - |
| F8 | -3.407E+03 | -1.164E+02 | - |
| F9 | 2.552E+01 | 1.519E+01 | + |
| F10 | 9.499E+00 | 8.926E-01 | + |
| F11 | 7.720E+00 | 1.307E-01 | + |
| F12 | 1.859E+03 | 3.723E-01 | + |
| F13 | 6.805E+04 | 9.926E-01 | + |

### 6.3.1 Solving a Real-World Problem using the MBA

In general, optimization depends on finding the best solution for a function to minimize or maximize a particular area. Many problems business has been solved using a variety of different approaches. Optimization problems are important concerning the business, industry, and even our daily lives. Literature has provided different solutions for scheduling problems, job assignments, call centers, task distribution, and many other optimization-related problems. This is done by showing how different types of algorithms work. These problems can be solved using linear programming. However, this programming is time-consuming and sometimes requires the use of many different mathematical equations. This is particularly true for exponentially complex problems. When faced with a small problem, it is possible to solve it quickly, but whenever there is a problem that grows in complexity exponentially, finding a solution quickly becomes problematic. Here is an example of using the scheduling problem in a restaurant. The problem was solved using a linear programming technique.

**Table 9: The restaurant example**

|  | Worker 1 | Worker 2 | Worker 3 | Worker 4 |
|---|---|---|---|---|
| **Job1** | 216 | 247 | 541 | 222 |
| **Job2** | 437 | 937 | 849 | 543 |
| **Job3** | 82 | 329 | 325 | 289 |
| **Job4** | 578 | 264 | 776 | 158 |

In this example, if job 1 is assigned to worker 3, then the expected work time will be 541 seconds. Also, if job 2 is assigned to worker 2, then the expected work time will be 937 seconds. The maximum solution from the matrix above is (541+937+289+578= 2345 seconds). This will require close to 9 minutes and 46 seconds of average work time. Most of the solutions for the above matrix will take 7 minutes on average. However, our primary goal is to find the minimum amount of time necessary to solve the problem. Using linear programming the minimum solution is (247+437+325+158=1157 seconds) or 4 minutes and 52 seconds. This result is a significant improvement in the average result of 7 minutes. The above example is solved with linear programming. A linear program uses a complex mathematical equation to solve the optimization problem and to get the final minimum solution.

This problem is a well-known one. The restaurant scheduling problem aims to allocate or distribute work and jobs in the most efficient way possible. Each worker has specific skills and experience, such as cooking, servicing, cleaning, moving quickly, etc.  At the same time, each job needs to be assigned to the right worker so that the best solution is found. For example, if there are four workers and four jobs, the requirement is to allocate the jobs to these workers. Logically, (4! =24) possible combinations exist, one of which is the optimal solution. In much the same way, a restaurant with 5 workers would have (5!=120) potential combinations. If the restaurant were to have ten workers with ten jobs, then (10! = 3,628,800) combinations would exist. Here the problem of using another type of programming, such as linear programming, occurs, because it is impossible to solve this number linearly and look through all of the combinations to find the optimal solution quickly.

The MBA based on the swarm intelligence algorithm is implemented to solve the above problem of restaurant scheduling. Every restaurant has many jobs and workers. The number of workers and jobs should be equal, which represents one dimension of the algorithm. In this application, there are 4 workers and 4 jobs. After running the application, the optimal solution for the restaurant job scheduling problem is 4.3734 minutes. Table (10) demonstrates the results of the application. The application is implemented with different dimensions starting from four to eight.

**Table 10: Result of application.**

| Dimension | Average Value | Standard Deviation | Work time in minutes |
|---|---|---|---|
| 4 | 266.0409 | 64.2047 | 4.434 |
| 5 | 304.7108 | 50.3414 | 5.0785 |
| 6 | 315.3420 | 59.0484 | 5.2557 |
| 7 | 337.8279 | 47.6988 | 5.6305 |
| 8 | 344.9468 | 55.3843 | 5.7491 |

The results shown in Table 10 indicate that the modified bat algorithm provides an optimal solution for the problem effectively and with minimal computation time.

## 7. Conclusion

In this paper, the BA and the proposed MBA were evaluated against 23 benchmark functions. The average value, standard deviation, and P-value results were calculated and assessed. Furthermore, the MBA was compared to the GA and the PSO with six functions. Also, further comparison between MBA and GA was conducted using unimodal and multimodal functions. As a result, the following conclusions were reached:

1) The BA is poor at exploration, whereas the MBA is more global and converges on results more quickly as it performed well on unimodal testing functions. Also, for the multimodal and fixed-dimension multimodal functions, the results of the MBA were significant since their optima values increased.
2) The average values of the MBA were better for 20 functions. The standard deviation results of the MBA were better for 17 functions.
3) The comparison between the MBA and the GA and PSO showed that the MBA performed better on all of the functions.
4) The comparison between the MBA and the GA showed that the MBA produced better results on eleven out of thirteen functions.